# Federated Learning Versus Classical Machine Learning: A Convergence Comparison

Muhammad Asad, Ahmed Moustafa, and Takayuki Ito
Department of Computer Science
Nagoya Institute of Technology, 466-8555, Nagoya - Japan
Email: a.muhammad.799@stn.nitech.ac.jp

*Abstract* — In the past few decades, machine learning has revolutionized data processing for large scale applications. Simultaneously, increasing privacy threats in trending applications led to the redesign of classical data training models. In particular, classical machine learning involves centralized data training, where the data is gathered, and the entire training process executes at the central server. Despite significant convergence, this training involves several privacy threats on participants' data when shared with the central cloud server. To this end, federated learning has achieved significant importance over distributed data training. In particular, the federated learning allows participants to collaboratively train the local models on local data without revealing their sensitive information to the central cloud server. In this paper, we perform a convergence comparison between classical machine learning and federated learning on two publicly available datasets, namely, logistic-regression-MNIST dataset and image-classification-CIFAR-10 dataset. The simulation results demonstrate that federated learning achieves higher convergence within limited communication rounds while maintaining participants' anonymity. We hope that this research will show the benefits and help federated learning to be implemented widely.

*Index Terms — Machine Learning; Federated Learning; Convergence; Artificial Intelligence;*

## I. INTRODUCTION

Artificial Intelligence and Machine Learning offer the ability to learn from experience and refine models without being specifically programmed [1]. These innovations have become common use in recent years: they have been used in several different fields and various applications [2]. For example, analysis of participant's data from a vehicle fleet in the automotive industry provides insights for customer needs, vehicle activities and driving environments [3]. New vehicle data analysis strategies include sending compressed sensor data from all vehicles to a central server and conducting the data analysis operations [4]. However, a modern car can generate hundreds of gigabytes of data in a single day, which means the data transfer and the storage of this data are virtually impossible [5]. Also, participants' data contain private information, which poses questions about the transmission and storage of these data on a central server [6].

To this end, a new approach called Federated Learning has been introduced, where only learning parameters of Deep Neural Network (DNN) are required to be communicated between the central server and the participants. At the same time, the whole training process is executed collaboratively on the individual participants [7]. This technique reduces the amount of data transfer and minimizes the privacy concerns of individual's private information. Hence, the load on powerful central servers in traditional machine learning has been distributed among individual low-power participants such as mobile devices or vehicles. Therefore, we can say that federated learning has the potential to challenge the dominant paradigm of distributed computation [8].

Federated learning varies in many ways from traditional machine learning problems (e.g., distribution of data centres). While both approaches strive to optimize their learning goal, federated learning algorithms have to consider the reality that contact with edge devices occurs over unstable networks with limited upload bandwidth [9]. As the communication overhead in federated learning is more often than computation overhead, therefore minimizing the communication overhead is crucial. This communication overhead can be measured either by uploading the gradients or through the communication rounds between the central server and participants [10]. The performance in federated learning can define with the achieved classification accuracy after specific numbers of communication rounds.

| Reference | Architecture | Model | Privacy Mechanism |
|---|---|---|---|
| [11] | Distributed | Decision Tree | Differential Privacy |
| [12] | Centralized | Decision Tree | Homomorphic Encryption |
| [13] | Centralized | Neural Network | Differential Privacy |
| [14] | Distributed | Decision Tree | Hashing |
| [15] | Centralized | Linear and NN Model | Homomorphic Encryption and Differential Privacy |
| [16] | Centralized | Linear Model | Homomorphic Encryption |

TABLE I: Existing approaches and their implications.

In similar ways, federated learning approaches varies between existing machine learning approaches such as; existing approaches make fundamental assumptions for the data training, which are much more robust in federated learning [17]. Below are the common assumptions made during the execution of traditional machine learning algorithms.

1. Data on the participants are sampled as independent and identically distributed (*i.i.d*). Whereas, federated learning assumes *non-i.i.d* as different users contain different types of data.

2. Data is evenly distributed among all the participants. This assumption is technically impossible, as the expected number and the actual number of participants are different in real-time scenarios. Therefore, federated learning divides the number of shards among the participated participants so that each participant can receive an equal amount of data.

3. Total number of participants is smaller than the available local training examples per participant. This assumption cannot be made in federated learning as federated learning is designed for large scale scenarios where the participants can be higher in number.

To this end, we conduct convergence comparison among classical machine learning and federated learning on a logistic-regression-MNIST dataset and image-classification-

CIFAR-10 dataset. In particular, we made the convergence comparison of classical centralized and distributed machine learning with general federated learning settings.

The rest of the paper is organized as follows: In Section II, we explain the brief background of machine learning and federated learning algorithms. In Section III, we differentiate the classifier models of machine learning and federated learning. In Section IV, we conduct simulation experiments and compare machine learning and federated learning. Finally, we conclude this paper in Section V.

## II. RELATED WORK

In this section, we provide extensive details of existing literature on machine learning and federated learning algorithms. As we have described in Section I, that classical machine learning faces the privacy threats of an individual's private information. Therefore, in this section, our primary focus is on privacy-preserving algorithms that proposed in the literature. The federated learning approach in data training phase has an advantage over classical machine learning, especially regarding the privacy preservation of an individual's data. For example, federated learning follows the concept of data minimization of the General Data Protection Regulation (GDPR), as only the trained model and no raw data are centrally processed. Communicated models are often transient because they are instantly discarded after integrating into the global model, which is an implementation of the GDPR concepts of storage and function limitation.

During the literature review, we observe that even with federated learning's unique features, this paradigm has not been widely implemented yet, and only a few articles have been proposed, where the major focus is based on communication and privacy concerns. Hence, various research gaps are still need to discover in federated learning, such as; heterogeneity of participants, clustering technique and the dynamic association of participants with clusters, hierarchical architecture, etc. [18].

In the literature, most of the work has been done on privacy-preserving algorithms for centralized and distributed learning algorithms. In [19], the authors conclude the significant challenges in machine-federated learning. In particular, they discussed data isolation and techniques for improving the data safety and privacy. To this end, the authors introduce three different architectures of federated learning; horizontal federated learning, vertical federated learning and federated transfer learning. Afterward, most of the literature on federated learning has been done on horizontal classification; therefore, vertical classification and federated transfer learning need special consideration to explore in the future.

On the one hand, in [20], the authors define the intersection between machine learning and privacy threats. To this end, the authors proposed an efficient way to protect the individual's data. In contrast, this technique requires data to be uploaded at the central server, which contradicts the privacy concerns raised by federated learning. Similarly, in [21], the authors launched several attacks on the machine learning platform and proposed the countermeasures to protect them. In particular, the authors briefly explained the threat model for machine learning and proposed the desired properties to improve the security and privacy threats. In [22], the authors provide an analysis of machine learning algorithms and their implications for an individual's concern in profiling the European Union General Data Protection Regulation (GDPR). In particular, the authors explored the rights of an individual's data protection and obligations, how the individual's data is collected and used and the consequences of machine learning algorithms on securing the data are also discussed.

On the other hand, in [23], the authors proposed sketched updates in federated learning using the large corpus of decentralized training data to minimize the communication overhead while maintaining privacy-preserving in the network. In [24], the authors proposed a multi-task federated learning framework that enables several ways to perform the required learning task by sharing the information and maintains the required safety threats. In [25], the authors proposed an efficient and secure aggregation of large-scale data. In particular, the authors proposed a secure multi-party computation to individual participants, where the participants compute the local updates and forward the parameters. The proposed framework can be implemented either in distributed machine learning or in federated learning by aggregating the individual's parameters of DNN. In addition to these frameworks, various research in the literature on cryptography techniques has also been conducted, such as differential privacy, homomorphic encryption and hashing technique.

In Table I, we summarize the existing literature of federated learning on cryptography techniques. In particular, we show the existing techniques on cryptography concerning their architecture, model, and mechanism where most of the work has been done on federated learning's centralized architecture. Hence, there is a lot of research gap on the distributed architecture of federated learning, where the participants train their models independently without the involvement of a centralized server.

## III. MODEL CLASSIFIERS

This section briefly explains the difference between centralized machine learning classifier, distributed machine learning classifier, and the federated learning classifier.

In any machine learning, the implementation can be done by following four stages; data pre-processing, network architecture, model training and model testing, respectively [26]. These stages are defined as follows:

1. Data pre-processing: The first task is to identify the learning task and declare the learning parameters for the pre-processing. Afterward, datasets are loaded to perform the training and import the required libraries, according to the learning task and datasets.

2. Network architecture: Once the data is pre-processed, it is loaded into the network architecture. The network architecture is chosen based on the learning task and the data. For example, for image classification, CNN performs better than MLP, whereas, for tabular data, MLP is preferred.

3. Model training: Afterward, the data is trained on the given network model using the machine learning framework such as TensorFlow, Keras, etc. The training involves several communication rounds to train the data.

4. Model testing: Once the training is done, the model is tested based on the achieved convergence accuracy.

In Figure 1, we show the model classifiers concerning the four stages mentioned above.

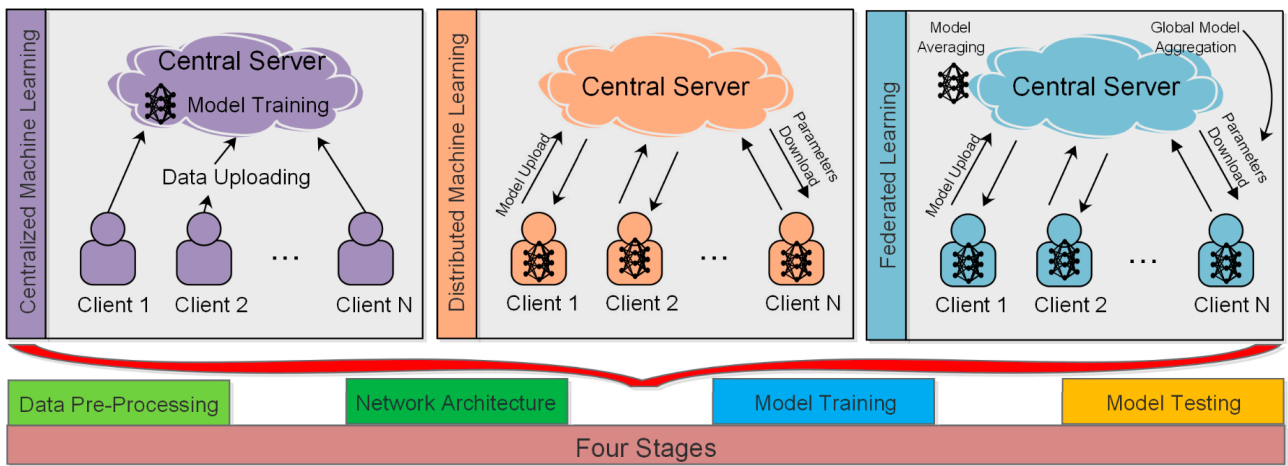

Fig. 1: Model Classifier of Centralized, Distributed and Federated Learning

## A. Centralized Machine Learning Classifier

In centralized machine learning, the participants are connected with the central server to upload their data. In particular, the participants upload their local data to the cloud server, and the cloud server performs all the computational tasks to train the data. On the one hand, centralized training is computationally-efficient for the participants as the participants are free from the computation responsibilities, which require higher resources. On the other hand, participants' private data is highly at risk as the cloud server can be malicious or infer through adversaries. Meanwhile, uploading a high range of data can also create communication overhead between participants and the cloud server.

## B. Distributed Machine Learning Classifier

Distributed machine learning algorithms are designed to resolve the computational problems of complex algorithms on large scale datasets. The distributed machine learning algorithms are more efficient and scalable than centralized algorithms. The distributed models are trained with same methodology as in centralized machine learning models, except they are trained separately on multiple participants [27]. During the training in a distributed algorithm, the participants independently train their models and send the weight updates to the central server. At the same time, the central server receives updates from participants and performs averaging for output. After the certain communication rounds, the convergence testing is done on the central cloud server.

## C. Federated Learning Classifier

Similar to distributed machine learning, federated learning also train the models independently. The only difference between distributed machine learning and federated learning is that in federated learning, each participant initializes the training independently as there is no other participant in the network. In federated learning, the training is conducted collaboratively and independently on individual participants. In particular, local epochs are declared in the learning parameters and each participant train the data by running the local epochs. After specific amounts of epochs, the local update is computed, and the participants send the updates to the cloud server. The cloud server receives the update from each participant, average them and aggregate the next global model. Based on this global model, the participants execute the training process for the next communication round. The process kept repeating until the desired convergence level is achieved or the given communication rounds are complete [28].

## IV. EXPERIMENTS AND RESULTS

This section conducts the experiments through simulations and achieves the comparative graphs among classical machine learning and federated learning.

### A. Experimental Setup

In the experiments, we use two open-source datasets, which are commonly used for federated learning experiments, namely; MNIST and CIFAR-10. We train these datasets on a convolution neural network (CNN) with 3 x 3 convolution layers and an output layer. The MNIST dataset has 60,000 training samples of handwritten images with a size of 28 x 28 pixels. On the other hand, the CIFAR-10 dataset has 50,000 training samples of images with a size of 32 x 32 pixels. The experiments are conducted on the server with an Intel(R) Core(TM) i9-9980HK CPU @ 2.40GHz and 32 GB of RAM.

### B. Results

In this subsection, we provide the details of our experiments and compare the classical machine learning and federated learning. In particular, we consider centralized machine learning, distributed machine learning and federated learning for comparison. To evaluate the results, we consider three different scenarios.

- Scenario one: We deploy *50* participants and set the *20%* participation ratio and simulate the centralized machine learning, distributed machine learning and federated learning algorithms, respectively for *100* communication rounds on MNIST and CIFAR-10 dataset.

- Scenario two: Similar to scenario one, here we consider the same parameters and execute all three algorithms, respectively, for *200* communication rounds on both datasets.

- Scenario three: In this scenario, we measure the convergence with the influence of participants. In particular, we deploy *p = {20, 40, 60, 80, 100}* participants and set the *20%* participation ratio and execute all three algorithms for *100* communication rounds on both MNIST and CIFAR-10 datasets.

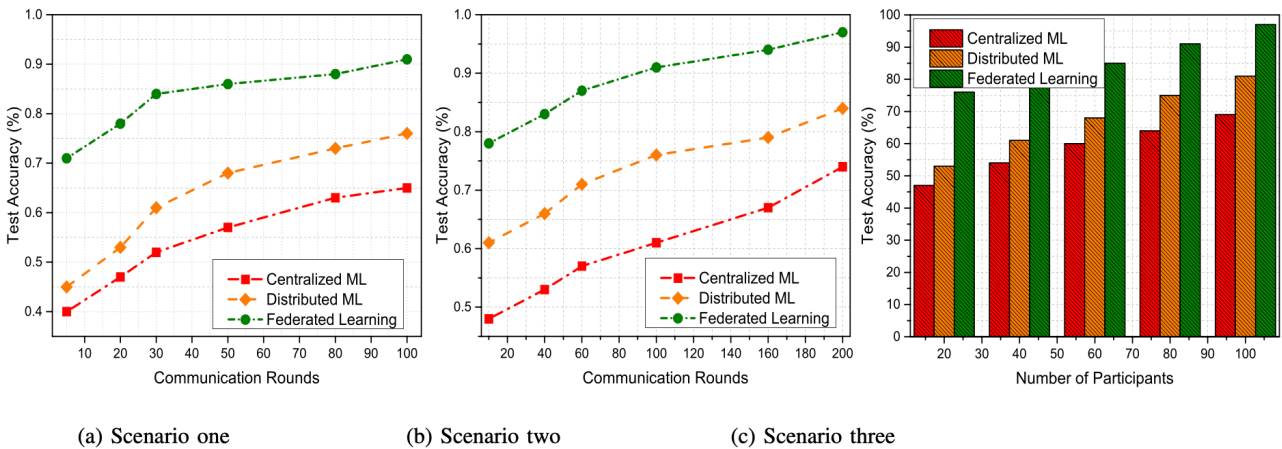

(a) Scenario one  (b) Scenario two  (c) Scenario three

Fig. 2: Convergence comparison on MNIST dataset in three different scenarios: 1) 100 communication rounds, 2) 200 communication rounds, 3) various number of participants.

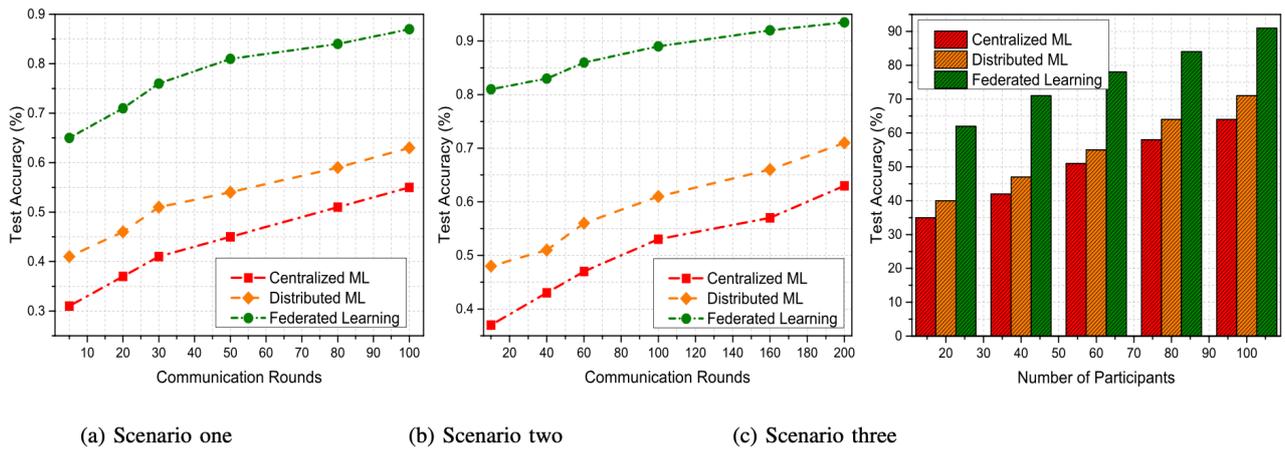

Fig. 3: Convergence comparison on CIFAR-10 dataset in three different scenarios: 1) 100 communication rounds, 2) 200 communication rounds, 3) various number of participants.

Figures 2 and 3 show the convergence comparison of centralized machine learning, distributed machine learning, and federated learning on MNIST and CIFAR-10 dataset on those mentioned above three different scenarios. The local epochs for training in distributed machine learning and federated learning are set to *10* for all the experiments. The graphs in Figures 2 and 3 show that the independent training achieves better performance than centralized machine learning.

The centralized machine learning suffers from the worst performance due to the large-scale data which is required to be uploaded at the central server in each communication round, where the participants have limited bandwidth capacity. In particular, centralized machine learning achieves only *65%* and *73%* of accuracy on the MNIST dataset for scenario one and scenario two, respectively. Similarly, centralized machine learning performs more poorly on the CIFAR-10 dataset and achieves only *54%* and *62%* of accuracy for scenario one and scenario two, respectively.

The distributed machine learning performs worse than federated learning but better than centralized machine learning. The fact behind this better performance is because the clients are independent of training their models. However, due to the privacy concern and involvement of the central server, distributed machine learning could achieve only *72%* and *78%* of accuracy in scenario one and scenario two on the MNIST dataset. Similar to the centralized, distributed machine learning also perform poor on CIFAR-10 datasets. The achieved accuracy of distributed machine learning on CIFAR-10 is only *67%* and *72%*, respectively.

Figure 2 and Figure 3 show that federated learning performs best and achieves the highest accuracy on both datasets. In particular, federated learning achieves *92%* and *97%* of accuracy on the MNIST dataset and *86%* and *94%* of accuracy on CIFAR-10 datasets on scenario one scenario two, respectively.

For the third scenario, all the machine learning algorithms suffer from the convergence when the participants are lower in number on both MNIST and CIFAR-10 datasets. As the dataset contains huge bits of information and the participants are limited for the training, therefore, the algorithms face a major computation bottleneck and perform worse. It is shown that when the number of participants increases, the training also gets better and face less computation overhead; hence, the performance on both datasets is linearly increasing with the increasing number of participants. In addition, in both datasets, federated learning performs best even with the limited number of participants.

Based on the experiments above, we can say that federated learning is the best possible solution in today's era. As malicious activities on social media and online platforms increase, day-by-day and traditional machine learning do not provide any security and privacy guarantee. Therefore, federated learning should be considered in all AI applications, especially where huge data and participants' private information are involved.

*V. CONCLUSION*

This paper compares the three different approaches of machine learning, namely; centralized machine learning, distributed machine learning, and federated learning. In the experiments, we simulated all three approaches on three different scenarios, where the federated learning outperform classical machine learning due to distributed and collaborative training while providing security features. Considering the achieved performance and increasing malicious activities on online platforms in today's era, we conclude that federated learning is a suitable platform for all AI applications, especially where massive data and an individual's private information is involved.

ACKNOWLEDGEMENT

This work has been supported by Grant-in-Aid for Scientific Research [KAKENHI Young Researcher] Grant No. 20K19931.

REFERENCES

1. S. Das, A. Dey, A. Pal, and N. Roy, "Applications of artificial intelli- gence in machine learning: review and prospect," *International Journal of Computer Applications*, vol. 115, no. 9, 2015.
2. I. M. Cockburn, R. Henderson, and S. Stern, "The impact of artificial intelligence on innovation," National bureau of economic research, Tech. Rep., 2018.
3. H. Fujiyoshi, T. Hirakawa, and T. Yamashita, "Deep learning-based image recognition for autonomous driving," *IATSS research*, vol. 43, no. 4, pp. 244–252, 2019.
4. S. Shalev-Shwartz, S. Shammah, and A. Shashua, "Safe, multi- agent, reinforcement learning for autonomous driving," *arXiv preprint arXiv:1610.03295*, 2016.
5. M. Patra, R. Thakur, and C. S. R. Murthy, "Improving delay and energy efficiency of vehicular networks using mobile femto access points," *IEEE Transactions on vehicular Technology*, vol. 66, no. 2, pp. 1496– 1505, 2016.
6. T. N. D. Pham and C. K. Yeo, "Adaptive trust and privacy management framework for vehicular networks," *Vehicular Communications*, vol. 13, pp. 1–12, 2018.
7. B. McMahan, E. Moore, D. Ramage, S. Hampson, and B. A. y Arcas, "Communication-efficient learning of deep networks from decentralized data," in *Artificial Intelligence and Statistics*. PMLR, 2017, pp. 1273– 1282.
8. S. Samarakoon, M. Bennis, W. Saad, and M. Debbah, "Distributed fed- erated learning for ultra-reliable low-latency vehicular communications," *IEEE Transactions on Communications*, vol. 68, no. 2, pp. 1146–1159, 2019.
9. J. Konečný, H. B. McMahan, F. X. Yu, P. Richtárik, A. T. Suresh, and D. Bacon, "Federated learning: Strategies for improving communication efficiency," *arXiv preprint arXiv:1610.05492*, 2016.
10. K. Yang, T. Jiang, Y. Shi, and Z. Ding, "Federated learning via over- the-air computation," *IEEE Transactions on Wireless Communications*, vol. 19, no. 3, pp. 2022–2035, 2020.

11. L. Zhao, L. Ni, S. Hu, Y. Chen, P. Zhou, F. Xiao, and L. Wu, "Inprivate digging: Enabling tree-based distributed data mining with differential privacy," in *IEEE INFOCOM 2018-IEEE Conference on Computer Communications*. IEEE, 2018, pp. 2087–2095.

12. Y. Liu, Z. Ma, X. Liu, S. Ma, S. Nepal, and R. Deng, "Boosting privately: Privacy-preserving federated extreme boosting for mobile crowdsensing," *arXiv preprint arXiv:1907.10218*, 2019.

13. A. Bhowmick, J. Duchi, J. Freudiger, G. Kapoor, and R. Rogers, "Pro- tection against reconstruction and its applications in private federated learning," *arXiv preprint arXiv:1812.00984*, 2018.

14. Q. Li, Z. Wen, and B. He, "Practical federated gradient boosting decision trees." in *AAAI*, 2020, pp. 4642–4649.

15. M. Asad, A. Moustafa, and T. Ito, "Fedopt: Towards communication efficiency and privacy preservation in federated learning," *Applied Sci- ences*, vol. 10, no. 8, p. 2864, 2020.

16. V. Nikolaenko, U. Weinsberg, S. Ioannidis, M. Joye, D. Boneh, and N. Taft, "Privacy-preserving ridge regression on hundreds of millions of records," in *2013 IEEE Symposium on Security and Privacy*. IEEE, 2013, pp. 334–348.

17. P. Kairouz, H. B. McMahan, B. Avent, A. Bellet, M. Bennis, A. N. Bhagoji, K. Bonawitz, Z. Charles, G. Cormode, R. Cummings *et al.*, "Advances and open problems in federated learning," *arXiv preprint arXiv:1912.04977*, 2019.

18. L. Lyu, H. Yu, and Q. Yang, "Threats to federated learning: A survey," *arXiv preprint arXiv:2003.02133*, 2020.

19. Q. Yang, Y. Liu, T. Chen, and Y. Tong, "Federated machine learning: Concept and applications," *ACM Transactions on Intelligent Systems and Technology (TIST)*, vol. 10, no. 2, pp. 1–19, 2019.

20. M. Al-Rubaie and J. M. Chang, "Privacy-preserving machine learning: Threats and solutions," *IEEE Security & Privacy*, vol. 17, no. 2, pp. 49–58, 2019.

21. N. Papernot, P. McDaniel, A. Sinha, and M. Wellman, "Towards the science of security and privacy in machine learning," *arXiv preprint arXiv:1611.03814*, 2016.

22. D. Kamarinou, C. Millard, and J. Singh, "Machine learning with personal data," *Queen Mary School of Law Legal Studies Research Paper*, no. 247, 2016.

23. K. Bonawitz, H. Eichner, W. Grieskamp, D. Huba, A. Ingerman, V. Ivanov, C. Kiddon, J. Konečný, S. Mazzocchi, H. B. McMahan *et al.*, "Towards federated learning at scale: System design," *arXiv preprint arXiv:1902.01046*, 2019.

24. V. Smith, C.-K. Chiang, M. Sanjabi, and A. S. Talwalkar, "Federated multi-task learning," in *Advances in Neural Information Processing Systems*, 2017, pp. 4424–4434.

25. K. Bonawitz, V. Ivanov, B. Kreuter, A. Marcedone, H. B. McMahan, S. Patel, D. Ramage, A. Segal, and K. Seth, "Practical secure aggregation for privacy-preserving machine learning," in *Proceedings of the 2017 ACM SIGSAC Conference on Computer and Communications Security*, 2017, pp. 1175–1191.

26. G. Forman and H. J. Suermondt, "Retraining a machine-learning classi- fier using re-labeled training samples," Sep. 7 2010, uS Patent 7,792,353. [27]

27. T. Kraska, A. Talwalkar, J. C. Duchi, R. Griffith, M. J. Franklin, and M. I. Jordan, "Mlbase: A distributed machine-learning system." in *Cidr*, vol. 1, 2013, pp. 2–1.

28. M. Duan, D. Liu, X. Chen, Y. Tan, J. Ren, L. Qiao, and L. Liang, "Astraea: Self-balancing federated learning for improving classification accuracy of mobile deep learning applications," in *2019 IEEE 37th International Conference on Computer Design (ICCD)*. IEEE, 2019, pp. 246–254.